# Testing, Verification and Improvements of Timeliness in ROS processes [1]

Mohammed Y. Hazim, Hongyang Qu, and Sandor M. Veres

Dept. of Automatic Control and Systems Engineering, University of Sheffield, UK
{myhazim1,h.qu,s.veres}@shef.ac.uk

**Abstract.** This paper addresses the problem of improving response times of robots implemented in the Robotic Operating System (ROS) using formal verification of computational-time feasibility. In order to verify the real time behaviour of a robot under uncertain signal processing times, methods of formal verification of timeliness properties are proposed for data flows in a ROS-based control system using Probabilistic Timed Programs (PTPs). To calculate the probability of success under certain time limits, and to demonstrate the strength of our approach, a case study is implemented for a robotic agent in terms of operational times verification using the PRISM model checker, which points to possible enhancements to the operation of the robotic agent.

## 1 Introduction

The Robot Operating System (ROS [10]) is an open-source operating system used to develop control software for robots. It has become popular due to its capabilities in perception, object detection, navigation, etc. and the increasing demand for a uniform platform for programmable robots. The correctness of a ROS program then attracts serious attention as the deployment of ROS grows rapidly. An important way to guarantee correctness in software is formal verification and several attempts have been conducted to apply it to ROS programs, such as [9][4][11]. ROSRV in [4] is a runtime verification framework on top of ROS in order to address safety and security issues of robots. The work in [9] considered the problem of generating a platform-specific glue code for platform-independent controller code in ROS, and the code generation process is amenable to formal verification. In [11], formal verification was applied to a high-level planner/scheduler for autonomous personal robotic assistants (Care-O-bot). However, none of the attempts addresses the performance alongside the correctness of a ROS program via formal verification to ensure stringent constraints on timeliness and other properties in ROS programs. This assurance is crucial to correct system behaviour and uncertainty in their environment.

This work is concerned with methods which can improve the performance of ROS based robot control systems. One of the difficulties in robot programming is to ensure that the robot responds to environmental challenges in a timely manner, let it be a threat approaching, to avoid something or the execution of a command which should not be delayed. Physical actions make the robot primarily depend on suitable speed of sensor signal processing, e.g., recognition and interpretation of relationships of static and moving objects in the environment, making sense of a command issued by a trusted human based on the context the robot and the human share, or planning of an action sequence to achieve a goal in a timely manner which does not render the goal outdated by the time the plan is ready, etc.

---

1    This work was supported by the EPSRC project EP/J011894/2.

The above computational challenges are addressed in the computational processes of ROS while a number of nodes are running, each in possibly several threads that communicate with each other between nodes. Broadcast of topics often interrupts subscriber nodes, and services requested from other nodes need to be waited for in order to be able to make use the data returned. For instance, sensing and recognition by computer vision may require some fixed or variable time, depending on the significant number of objects of the environment. Discovering relationships in the environment may however take even more variable time to compute. Clearly action taking can suffer delays as planning cannot start before relationships are modelled. We propose that improvements to ROS-based computational performance can be analysed and carried out in three phases:

1. Statistical modelling of computational times in various categories and complexities of perception (including sensing and analysis), planning and execution of planned actions.
2. Formal analysis of the statistically modelled given ROS system using probabilistic timed programs (PTPs) [2] by answering PCTL queries on unacceptable delays in computation in operations by model checker PRISM [7].
3. Revision of procedures used in the ROS system to reduce the chance of computational delays.

In this work, we first design a ROS system in a rational agent framework LISA (Limited Instruction Set Architecture) [5], which is based on AgentSpeak expansions such as Jason and Jade, with more focusing on external planning process, abstraction from planning and optimisation from decision making. The LISA model is then compiled into a PTP model for the formal analysis.

## 2 The Robot Operating System

ROS is not a traditional Operating System. Rather it provides a structured communications layer in which individual processes can interact [10]. It simplifies the task of programming robots by providing a robust framework where the designer is provided a declarative programming environment for parallel computational processes of a robot. A ROS implementation of a robotic software has three typical components:

- Nodes - Nodes are basic processes that perform the sensing, computation and control tasks. Typically, each node can contain several computational threads, although it may have additional sub-threads which the programmer is responsible for designing. Typical systems are formed from many nodes, each of which does a portion of the overall task.
- Services - Services provide a strict communication model where there is an established request and response message between two nodes. In a process similar to web services, a node may subscribe and subsequently request in- formation via a service and then be supplied back with the information on demand.
- Topics - In order to publish messages any node can establish a topic and publish messages to it, as and when necessary. Any other node within the network may also publish to this topic. In order to receive messages, the other nodes may subscribe, wherein they can receive any message sent via a call back. A topic is a broadcast messaging stream and so does not provide any synchronous message transfer.

A fundamental difference between services and topics is that services are re- quester/receiver initiated while topics are sender/provider initiated and the receivers are immediately notified, asynchronously. Both are however many-to-many communications as there can be several providers and receivers of any service or topic. Topics are inefficient when a node only needs some data from another node occasionally, when it needs it; while services are inefficient when a node needs some data supplied on a continuous, "as soon as possible" basis, though asynchronously. In their own way both are efficient ways to communicate for different purposes. Care needs to be taken however that a subscriber to a topic does not receive more data than it needs as otherwise it is wasting its computational resources on handling redundant messages from the topic. For instance, sensor messages are to be published to a topic only with a frequency which is needed by other nodes, thereby resulting in less latency than if a service were doing the same job.

## 3 Mathematical model of a ROS Package

One way to describe a ROS based system is a tri-partite graph with vertices for nodes, topics and services. These vertex types are not interchangeable in graph matching algorithms. New topics and services can be easily introduced that can allow reconfiguration of the system to provide agents with the information they required, albeit sourced from different locations. All node communication must occur through topics or services.

Definition 3.1. A ROS-graph is $G = (N, T, S, E, D, C, X, \lambda)$, where $N$ are the set of vertices representing ROS nodes, $T$ are a set of topics and $S$ are a set of services, $C$ is a partially order set of object classes and $X$ is a set of labels to name all vertices. $E \subset (N \times T) \cup (T \times N) \cup (N \times S) \cup (S \times N)$, is a set of directed edges to represent publishing of, and subscription to, topics and provision of, and subscription to, services, respectively. $D : E^- \to C^*, E^- = T \cup (N \times S) \cup (S \times N)$, is a data descriptor function where $C^*$ is a notation for finite sequences of entries from the set of data object classes $C$, which are used in services and topics to send information between nodes. Each of $N, T, S$ are labelled by a surjective labelling function $\lambda: N \cup T \cup S \rightarrow X$.

A ROS system enables the nodes to advertise or use services, and to publish or subscribe to topics. G represents the maximum ability of the robot when the system has all nodes, topics and services nominally functioning. If some nodes are not available due to sensor, actuator or computational hardware breakdown, then G needs sufficient redundancy to enable continued functioning of the robot or at least some of its functionality. The ROS graph G defines all the possible data flows for sensor readings, signal processing and control action in the environment. A detailed description is not within the scope of this work and we refer the reader to [1].

## 4 Statistics of ROS nodes

When ROS based robot control system's programming is completed, the robot is ready to be tested in a series of scenario tests. Performance may not acceptable due to a few factors:

1. When a plan of an agent is triggered due to environmental change the computational times of perception modelling and planning are excessive and delay action taking in some environmental scenarios.

2. In some environmental scenarios scene interpretation and planning is several times faster than typical response time requires. The question arises whether more complex model of the scene could have been built to more fully grasp an environmental situation.

Overall the performance problem of the robot is to discover scenarios which are not favourable for the robot's computational system. These are searched and synthesised based on sensor and perception statistics derived in practical use of the ROS system. This section provides a formal model of statistical estimation of computation and communication times in a given ROS system already operating on a hardware platform. Consequent application of probabilistic model checking can guide us to introduce improvements in the choice of computational processes involved in reasoning.

### 4.1 Performance evaluator node

To estimate the processing and communication time across the ROS system and additional runtime statistics node $\Sigma$ can be introduced, which collects runtime data from all the robots functional nodes. Each of the functional nodes $i$ has a data array $D_i$ recording timed-performance of services and topics in the node. Let denote $s_k \in S$ a service in a ROS-graph $G = (N, T, S, E, D, C, X, \lambda)$. The following timed data are recorded about a service call.

1. When a request is to be made from node $j$ for service $s_k$, then a data entry $(n_j \xrightarrow{\text{req}} s_k, t^j)$ is added to $D_j$ just before the service command is issued from node $j$ to node $i$ with time stamp $t^j$ in node $j$.

2. Upon request, and before any execution of service actions, a data entry $(n_j \xrightarrow{req} s_k, t^i)$ is added to $D_i$ with time stamp $t^i$ in node $i$.

3. Upon completion of the computational processes or physical controls performed, a data entry $(n_j \xrightarrow{ans} s_k, t^i)$ is added to $D_i$ with time stamp in node $i$.

4. Upon answer data received in node $j$ for service $s_k$, then a data entry $(n_j \xrightarrow{ans} s_k, t^j)$ is added to $D_j$ with time stamp $t^j$ in node $j$.

For topics recording of runtime data is slightly different:

1. When a topic is to be published by node $j$ for topic $p_k$, then a data entry $(n_j \xrightarrow{pub} p_k, t^j)$ is added to $D_j$ just before the topic boadcast is issued from node $j$ with time stamp $t^j$ in node $j$.

2. Upon receiving the broadcast, and before any execution of actions due to the topic broadcast, a data entry $(n_j \xrightarrow{rec} p_k, t^i)$ is added to $D_i$ with time stamp $t^i$ in node $i$.

3. Upon completion of the computational processes or physical controls performed, a data entry $(n_j \xrightarrow{top} s_k, t^i)$ is added to $D_i$ with time stamp in node $i$.

Note that there are other ways to collect statistics on execution time and latency, such as in [3], but our method suits our need better because it does not depend on the header of messages, which is not always available.

## 4.2 Estimation of operations

From each node $i$ the data containers $D_i$ are sent to the runtime statistics node $\Sigma$, which can compute the following amongst others:
- Probability distribution of the request communication times $t^i - t^j$ from $(n_j \xrightarrow{req} s_k, t^j)$ and $(n_j \xrightarrow{req} s_k, t^i)$.
- Probability distribution of the service execution times $t^{si} - t^{ei}$ f from $(n_j \xrightarrow{req} s_k, t^{si})$ and $(n_j \xrightarrow{ans} s_k, t^{ei})$.
- Probability distribution answer communications times $t^j - t^i$ from $(n_j \xrightarrow{ans} s_k, t^j)$ and $(n_j \xrightarrow{ans} s_k, t^i)$.
- Probability distribution of communication broadcast times $t^i - t^j$ from of $(n_j \xrightarrow{pub} p_k, t^j)$ and $(n_j \xrightarrow{rec} p_k, t^i)$.

- Probability distribution of topic interruption times, $t^{si} - t^{ei}$ from $(n_j \xrightarrow{rec} p_k, t^{si})$ and $(n_j \xrightarrow{top} s_k, t^{ei})$.

Performance tuning of a ROS based computational system is carried out iteratively through a series of trial runs, during which the average runtime probabilities (or conditional runtime probabilities of the duration events are evaluated), followed by a ROS system. This is followed by algorithmic adjustments made to the ROS system and the iteration continues by another trial run. The series of iterations consisting of (1) trial-run (2) compilation to PRISM model (3) running of PCTL queries (4) algorithmic amendments are cyclically repeated on the ROS system until satisfactory computational performance is achieved.

## 5 A rational agent framework LISA

Comparing with Jason in terms of plan selection function, LISA [5] proved to enhance the architecture with a runtime probabilistic model checking by predicting the outcomes of applicable plan and selections. The LISA structure is simpler than its predecessors and can easily lend itself to design time and run-time verification. Now we give the detail about LISA. By analogy to previous definitions [8, 12] of AgentSpeak-like architectures, we define our agents as a tuple: $\mathsf{R} = (\mathsf{F}, B, L, \Pi, A)$, where:

- $\mathsf{F} = \{p_1, p_2, \ldots, p_{n_p}\}$ is the set of all predicates.
- $B \subset \mathsf{F}$ is the total set of belief predicates. The current belief base at time $t$ is defined as $B_t \subset B$. Beliefs that are added, deleted or modified can be either called *internal* or *external* depending on whether they are generated from an internal action, in which case are referred to as "mental notes", or from an external input, in which case they are called "percepts".
- $L = \{l_1, l_2, \ldots l_{n_l}\}$ is a set of logic-based implication rules.
- $\Pi = \{\pi_1, \pi_2, \ldots, \pi_{n_\pi}\}$ is the set of executable plans or *plans library*.
  Current applicable plans at time $t$ are part of the subset applicable plan $\Pi_t \subset \Pi$ or "desire set".
- $A = \{a_1, a_2, \ldots, a_{n_a}\} \subset \mathsf{F} \setminus B$ is a set of all available actions. Actions can be either *internal*, when they modify the belief base or data in memory objects, or *external*, when they are linked to external functions that operate in the environment.

AgentSpeak like languages, including LISA, can be fully defined and implemented by specifying initial beliefs and actions, and reasoning cycles:

- *Initial Beliefs*. The initial beliefs and goals $B_0 \subset F$ are a set of literals that are automatically copied into the *belief base* $B_t$ (that is the set of current beliefs) when the agent mind is first run.

- *Initial Actions*. The initial actions $A_0 \subset A$ are a set of actions that are executed when the agent mind is first run. The actions are generally goals that activate specific plans.

The following operations are repeated for each reasoning cycle in AgentSpeak.

- *Maintenance of Percepts*. This means generation of perception predicates for $B_t$ and data objects such as the world model used here $W$.
- *Logic rules*. A set of logic based implication rules $L$ describes *theoretical* reasoning to improve the agent current knowledge about the world.
- *Executable plans*. A set of *executable plans* or *plan library* $\Pi$. Each plan $\pi_j$ is described in the form: $p_j : c_j \leftarrow a_1, a_2, \ldots, a_{n_j}$, where $p_j \in B$ is a *triggering predicate*, which allows the plan to be retrieved from the plan library whenever it comes true, $c_j \in B$ is a logic formula of a *context*, which allows the agent to check the state of the world, described by the current belief set $B_t$, before applying a particular plan sequence $a_1, a_2, \ldots, a_{n_j} \in A$ with a list of actions. Each $a_j$ can be one of (1) predicate of an external action with arguments of names of data objects, (2) internal (mental note) with a preceding + or - sign to indicate whether the predicate needs to be added or taken away from the belief set $B_t$ (3) conditional set of items from (1)-(2). The set of all triggers $p_j$ in a program is denoted by $E_{tr}$

LISA enhanced the above reasoning cycle to allow multiple actions to be executed in parallel. The enhanced reasoning cycle consists of the following steps:

1. *Belief base update*. The agent updates the belief base by retrieving information about the world through perception and communication. Adding and removing beliefs from the belief base is carried out by the function *Belief Update Function (BUF)*.
2. *Application of logic rules*. The logic rules in $L$ are applied in a round-robin fashion (restarting at the beginning of the list) until there are no new predicates generated for $B_t$. This means that rules need to be verified not to lead to infinite loops.
3. *Trigger Event Selection*. For every reasoning cycle a function called *Belief Review Function* $S_t : \wp(B_t) \rightarrow \wp(E_t)$ selects the current event set $E_t$, where $\wp(\cdot)$ is the so called *power operator* and represents the set of all possible subset of a particular set. We call the current selected trigger event $S_t(B) = T_t$ and the associated plans the *Intention Set*.
4. *Plan Selection*. All the plans in $T_t$ are checked for their context to form the *Applicable Plans* set $\Pi_t$ by function $S_O : E_t \rightarrow \wp(S_t)$. We will call the current selected plan $S_O(\Pi_t) = \pi_t$.

5. *Plan Executions*. All plans in $S_O : E_t$ are started to be executed concurrently by going through the plan items $a_1, a_2, \ldots, a_{n_j}$ one-by-one sequentially.

# 6 Modelling of agent operational times in PRISM

In this section we assume that the response of the physical environment of the agent is modelled as a probabilistic timed program (PTP) $E$ in terms of the predicates feed back to the belief base of the agent under various environmental states. $E$ is composed of environmental states, and transitions which under each state through the conditional probabilities of the environment corresponds to triggering of predicates through the sensor system of the robotic agent. Given that the agent has well defined decision structures as described in the previous section, the environment-agent model will also be a PTP. This section describes how the combination of probability distributions, which were estimated in the previous section, when combined with the environmental PTP and the logic based decision making of the agent, can be modelled in PRISM.

## 6.1 Probabilistic timed programs (PTP)

*Probabilistic timed programs* [6] are an extension of Markov Decision Processes (MDPs) with state variables and real-time clocks.

Given a set $\mathcal{V}$ of variables, let $(\mathcal{V})$, $Val(\mathcal{V})$ and $(\mathcal{V})$ be a set of *assertions*, *valuations* and *assignments* over $\mathcal{V}$ respectively. Given a set $S$, let $S$ be the set of subsets of $S$ and $S$ the set of discrete probability distributions over $S$. A set $\mathcal{X}$ of *clock* variables represents the time elapsed since the occurrence of various events. The set of *clock valuations* is $\mathbb{R}_{\geq 0}^{\mathcal{X}} = \{t : \to \mathbb{R}_{\geq 0}\}$. For any clock valuation $t$ and any $\delta \geq 0$, the *delayed* valuation $t + \delta$ is defined by $(t + \delta)(x) = t(x) + \delta$ for all $x \in \mathcal{X}$. For a subset $Y \subseteq \mathcal{X}$, the valuation $t[Y := 0]$ is obtained by setting all clocks in $Y$ to 0: $t[Y := 0](x)$ is 0 if $x \in Y$ and $t(x)$ otherwise. A (convex) *zone* is the set of clock valuations satisfying a number of clock difference constraints, i.e. a set of the form: $\rho = \{t \in \mathbb{R}_{\geq 0}^{\mathcal{X}_0} \mid t_i - t_j \lesssim b_{ij}\}$. The set of all zones is $Zones(\mathcal{X})$.

**Definition 1 (PTP).** A PTP is a tuple $P = (L, l_0, \mathcal{X}, \mathcal{V}, v_i, \mathcal{I}, \mathcal{T})$ where:

- $L$ is a finite set of *locations* and $l_0 \in L$ is the *initial location*;
- $\mathcal{X}$ is a finite set of *clocks* and $\mathcal{I} : S \to Zones(\mathcal{X})$ is the *invariant condition*;
- $\mathcal{V}$ is a finite set of *state variables* and $v_i \in Val(\mathcal{V})$ is the *initial valuation*;
- $\mathcal{T} : S \to Trans(L, \mathcal{V}, \mathcal{X})$ is the *probabilistic transition function*, where $Trans(L, \mathcal{V}, \mathcal{X}) = \text{Asrt}(\mathcal{V}) \times Zones(\mathcal{X}) \times \text{D}(\text{Assn}(\mathcal{V}) \times \text{P}(\mathcal{X}) \times L)$.

A step from a state $(l, v, t)$ consists of the elapse of a certain amount of time $\delta \in \mathbb{R}_{\geq 0}$ followed by a *transition* $\tau = (\mathcal{G}, \mathcal{E}, \Delta) \in \mathcal{T}(l)$. The transition comprises a *guard* $\mathcal{G} \in \text{Asrt}(\mathcal{V})$, *enabling condition* $\mathcal{E} \in Zones(\mathcal{X})$ and probability distribution $\Delta = \lambda_1(f_1, r_1, l_1) + \cdots +$ $\lambda_k(f_k, r_k, l_k))$ over triples containing an *update* $f_j \in \text{Asrt}(\mathcal{V})$, clock *resets* $r_j \subseteq \mathcal{X}$ and *target location* $l_j \in L$.

The delay $\delta$ must be chosen such that the invariant $\mathcal{I}(l)$ remains continuously satisfied; since $\mathcal{I}(l)$ is a (convex) zone, this is equivalent to requiring that both $t$ and $t + \delta$ satisfy $\mathcal{I}(l)$. The chosen transition $\tau$ must be *enabled*, i.e., the guard $\mathcal{G}$ and the enabling condition $\mathcal{E}$ in $\tau$ must be satisfied by $v$ and $t + \delta$, respectively. Once $\tau$ is chosen, an assignment, set of clocks to reset, and successor location are selected at random, according to the distribution $\Delta$ in $\tau$.

## 6.2 Performance queries

Given a PTP, we can use the following PCTL queries to check its properties:

- $P_{\bowtie=?}[F\ a]$,
- $P_{\bowtie=?}[F_{\leq T}\ a]$,

where $\bowtie\, \in \{max,\ min\}$, $a$ is Boolean expression that does not refer to any clocks and $T$ is an integer expression. The first query asks what is the maximum/minimum probability that $a$ is satisfied, and the second one inquires the probability that $a$ can be satisfied within time bound $T$. Based on these queries, we can compute the maximum/minimum probability of all target states that satisfy $a$ without time limit or within a bound $T$. For example, we can ask what is the minimum probability for a robot moving to a specific location within certain time. A concrete example will be shown in the next section.

## 6.3 Verification process

Figure 1 illustrates the whole process in our method. A system is first written in LISA and then translated into ROS. A performance evaluator node is generated for this system. After the evaluator node collects sufficient statistics on the time delay, it computes the probability distribution. A PTP model is then constructed using this information and the LISA program, although it is feasible to build the PTP model from the ROS program directly. The reason that we build the PTP model from the LISA program is that it provides a high level abstraction of the system, which can make the PTP model compact. The PTP model is fed to PRISM for verification. The result is then used as a reference when improving the design of the ROS program.

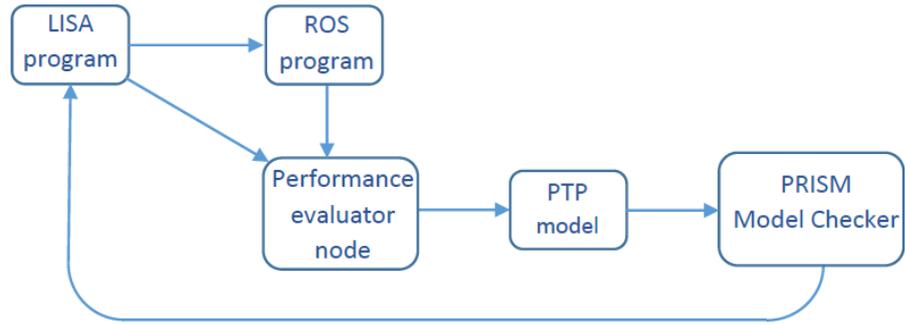

Figure 1: The verification process.

## 7 Case study

In this section we demonstrate the strength of our approach using the following scenario. An autonomous ground vehicle (AGV) is exploring a remote area with a vision system consisting two cameras (primary and secondary camera). The system merges two images, one from each camera, to look for an object in the area. Here we are mainly interested in two ROS nodes: one for receiving images from the cameras and the other for processing these images. The statistics shows that time for receiving one image respect the following probability distribution:

• With probability 0.3, it take less than 4 units of time, but more than 3 units to receive one image;

• With probability 0.6, the receiving time locates in the interval (4,6);

• With probability 0.1, the receiving time locates in (6,8).

It takes less than 16 units of time but more than 12 units to process two images, and the probability of successfully finding the object in the images is 0.91. When the system fails to find the object, it will take two new images from the cameras and repeat the process.

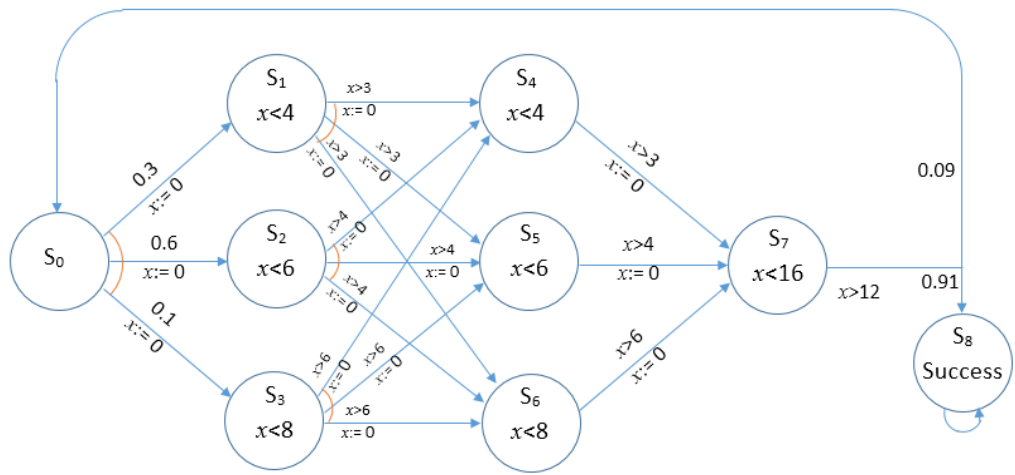

Figure 2: The PTP for the system.

Figure 2 illustrates the PTP model for the system, where $x$ is a clock, which is used to count the time elapse for each step. The timing constraints in a node (which represents a state), such as $x < 4$, is the upper bound and the constraints on an edge (which represents a transition), such as $x > 3$, is the lower bound. This figure shows that the system receives the image from the primary camera first (states $s_1$, $s_2$ and $s_3$), and then receives the one from the secondary camera (states $s_4$, $s_5$ and $s_6$). In state $s_7$, the system processes the images. We can ask a query that at what probability the system successfully find the object within 35 units of time, which can formulated in PCTL as follows:

$$P_{max=?}[F_{\leq 35} \ \text{``Success''}] \tag{1}$$

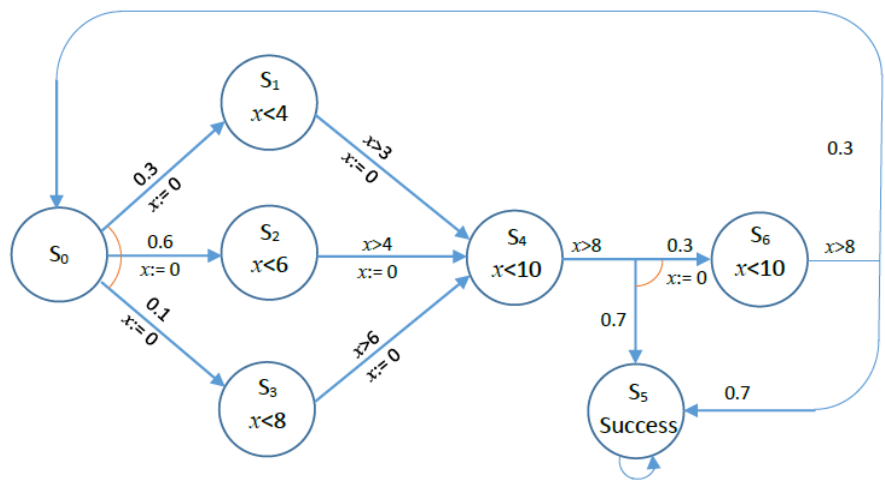

Figure 3: The PTP for the new system.

The result returned by PRISM is 0.91. One problem in this system is that it has to wait for two images before it can start to look for the object. If the image processing and receiving can be performed in parallel by different hardware, we may be able to increase the performance of the system, which is possible if the object can be found from one image, even if at a lower probability, e.g., 0.7. One way to achieve it as follows. The system starts to process the first image immediately after it arrives. Here we assume that processing one image is between 8 and 10 units. As the processing time exceeds the time required for receiving an image, the system does not need to wait once it finishes processing the first image. Instead, it can immediately process the second images. Although it is slightly slower to process the images separately than processing them altogether, eliminating the waiting time for the second image makes the system able to receive more images within the time limit and thus, find the object at higher probability. Figure 3 illustrates the improved system design. The result for the query in Equation (1) is 0.9724, which shows a big improvement from the previous design. Figure 4 shows the PRISM program for this improved system.

```
1  pta
2  module M
3    s : [0..6];
4    x : clock;
5    [] s=0 -> 0.3:(s'=1)&(x'=0) + 0.6:(s'=2)&(x'=0) +
          0.1:(s'=3)&(x'=0);
6    [] s=1 & x<4 & x>3 -> (s'=4)&(x'=0);
7    [] s=2 & x<6 & x>4 -> (s'=4)&(x'=0);
8    [] s=3 & x<8 & x>6 -> (s'=4)&(x'=0);
9    [] s=4 & x<10 & x>8 -> 0.7:(s'=5) + 0.3:(s'=6)&(x'=0);
10   [] s=5 -> (s'=5);
11   [] s=6 & x<10 & x>8 ->0.7:(s'=5)+0.3:(s'=0)&(x'=0);
12 endmodule
```

Figure 4: The Prism program for the new system.

## 8 Conclusions

This paper presented a method for formal verification of timeliness properties of robots implemented in ROS. The LISA framework was used to design a robotic agent as LISA provides a solution for the verification of robotic agents through the PRISM model checker. Statistical estimation was applied to robot operations under the ROS system to detect and collect information about the latency in the system. The LISA model was then associated with runtime probabilities and translated into a PTP model and verified in PRISM. It has been illustrated how to apply the methods to improve the design of a ROS system in a case study.

In the future we intend to bring the methods nearer to industrial applicability by improving their timing performance analysis, which might require the development of more efficient model checking algorithms for PTPs in the case of very large models. Another direction of future work is to search for other modelling formalisms, which can handle continuous probability distributions on timing variances, as PTP can only deal with discrete probability distributions.

## References

1. J. M Aitken, S. M. Veres, and M. Judge. Adaptation of system configuration under the robot operating system. In Proc. of the 19th world congress of the international federation of automatic control, 2014.
2. K. Drager, M. Z. Kwiatkowska, D. Parker, and H. Qu. Local abstraction refinement for probabilistic timed programs. Theor. Comput. Sci., 538:37-53, 2014.
3. D. Forouher, J. Hartmann, and E. Maehle. Data ow analysis in ROS. In Proc. Of the 41st International Symposium on Robotics, pages 1{6. VDE, 2014.
4. J. Huang, C. Erdogan, Y. Zhang, B. Moore, Q. Luo, A. Sundaresan, and G. Rosu. ROSRV: Runtime verification for robots. In Proc. of the 14th International Conference on Runtime Verification, LNCS 8734, pages 247{254. Springer, 2014.

5. P. Izzo, H. Qu, and S. M. Veres. Reducing complexity of autonomous control agents for verifiability. arXiv:1603.01202[cs.SY], March 2016.
6. M. Kwiatkowska, G. Norman, and D. Parker. A framework for verification of software with time and probabilities. In Proc. of the 8th International Conference on Formal Modelling and Analysis of Timed Systems, LNCS 6246, pages 25-45. Springer, 2010.
7. M. Kwiatkowska, G. Norman, and D. Parker. Prism 4.0: Verification of probabilistic real-time systems. In Computer aided verification, pages 585{591. Springer, 2011.
8. N. K. Lincoln and S. M. Veres. Natural language programming of complex robotic BDI agents. Intelligent and Robotic Systems, 71(2):211-230, 2013.
9. W. Meng, J. Park, O. Sokolsky, S. Weirich, and I. Lee. Verified ROS-based deployment of platform-independent control systems. In Proc. of NASA Formal Methods, LNCS 9058, pages 248-262. Springer, 2015.
10. M. Quigley, K. Conley, B. P. Gerkey, J. Faust, T. Foote, J. Leibs, R. Wheeler, and A. Y. Ng. ROS: an open-source Robot Operating System. In ICRA Workshop on Open Source Software, volume 3, 2009.
11. M. Webster, C. Dixon, M. Fisher, M. Salem, J. Saunders, K. Koay, and K. Dautenhahn. Formal verification of an autonomous personal robotic assistant, pages 74{79. AAAI, 2014.
12. M. Wooldridge. An Introduction to MultiAgent Systems. Wiley, Chichester, 2002.